\documentclass[letterpaper, 10 pt, conference]{ieeeconf}  % Comment this line out if you need a4paper

\IEEEoverridecommandlockouts                              % This command is only needed if 
\usepackage{graphics} % for pdf, bitmapped graphics files
\usepackage{epsfig} % for postscript graphics files
\usepackage{mathptmx} % assumes new font selection scheme installed
\usepackage{times} % assumes new font selection scheme installed
\usepackage{amsmath} % assumes amsmath package installed
\usepackage{amssymb}  % assumes amsmath package installed
\usepackage{makecell}
\usepackage{booktabs}
\usepackage{lipsum}
\usepackage{tikzducks}
\usepackage{siunitx}
\usepackage{xspace}
\usepackage[colorlinks=true, linkcolor=orange, urlcolor=orange, citecolor=orange]{hyperref}
\usepackage[font=footnotesize]{caption}
\usepackage{balance}

\DeclareMathAlphabet{\mathcal}{OMS}{cmsy}{m}{n}

\usepackage{xcolor}
\usepackage{colortbl}
\definecolor{mygray}{gray}{0.88}
\definecolor{myred}{HTML}{FF6666}
\definecolor{myblue}{HTML}{3399FF}
\definecolor{myorange}{HTML}{FFCC99}

\newcommand{\method}{ICLR\xspace}

\newcommand{\iclrurl}{%
  \href{https://toannguyen1904.github.io/ICLR}{%
    \textbf{\texttt{toannguyen1904.github.io/ICLR}}
  }%
}

\title{\LARGE \bf
\method: In-Context Imitation Learning with Visual Reasoning
}

\author{
    \textbf{Toan Nguyen}\textsuperscript{$\diamondsuit$} \quad
    \textbf{Weiduo Yuan}\textsuperscript{$\diamondsuit$} \quad
    \textbf{Songlin Wei}\textsuperscript{$\diamondsuit$} \quad
    \textbf{Hui Li}\textsuperscript{$\blacklozenge$}\\ 
    \textbf{Daniel Seita}\textsuperscript{$\diamondsuit$, $\ddagger$}\thanks{$^{\ddagger}$ Equal advising} \quad 
    \textbf{Yue Wang}\textsuperscript{$\diamondsuit$, $\ddagger$}\\ \\
    \textsuperscript{$\diamondsuit$}University of Southern California \quad
    \textsuperscript{$\blacklozenge$}Autodesk Research\\ \\
    \iclrurl
}

\begin{document}

\maketitle
\thispagestyle{empty}
\pagestyle{empty}

%%%%%%%%%%%%%%%%%%%%%%%%%%%%%%%%%%%%%%%%%%%%%%%%%%%%%%%%%%%%%%%%%%%%%%%%%%%%%%%%
\begin{abstract}
In-context imitation learning enables robots to adapt to new tasks from a small number of demonstrations without additional training. However, existing approaches typically condition only on state–action trajectories and lack explicit representations of task intent. This limitation hinders performance in complex and ambiguous task settings where the same actions may be consistent with different objectives. To address this, we present In-Context Imitation Learning with Visual Reasoning (\method), a novel framework that augments demonstration prompts with structured visual reasoning traces representing anticipated future robot trajectories in image space. \method also jointly learns to generate reasoning traces and low-level actions within a unified autoregressive transformer, enabling the model to mimic not only action prediction but also the reasoning process that leads to those actions. We extensively evaluate \method in both simulation and real-world manipulation tasks and demonstrate consistent improvements in success rates and generalization to unseen tasks and novel object configurations compared to other in-context imitation learning methods. These results suggest that incorporating embodied visual reasoning represents a promising direction for enhancing the robustness and generalization of robotic in-context learning systems.
% Our project website is available at: \href{https://toannguyen1904.github.io/roboreason}{https://toannguyen1904.github.io/roboreason}
% \yue{let's add results to the website as well}.
\end{abstract}

% \vspace{2ex}

%%%%%%%%%%%%%%%%%%%%%%%%%%%%%%%%%%%%%%%%%%%%%%%%%%%%%%%%%%%%%%%%%%%%%%%%%%%%%%%%
\section{INTRODUCTION}
A long-standing and significant challenge in robotics is data scarcity~\cite{goldberg2025good}. Collecting large-scale demonstration data for robotic manipulation in the real world is labor-intensive, time-consuming, and can pose several safety risks~\cite{khazatsky2024droid}. This has motivated the development of robot learning methods that can quickly acquire new skills from a limited number of robot demonstrations~\cite{jang2022bc,chen2025rovi,fu2025icrt,yang2025novel,ye2026world,lin2026systematic}. One promising direction that has recently attracted significant attention is in-context imitation learning~\cite{di2024keypoint,fu2025icrt,vosyliusinstant,shah2026mimicdroid,Jiang2026RoboTTTCS,zhang2026native}. In this learning paradigm, the robot learning model is trained to condition its behavior on a set of ``in-context,'' or ``prompt,'' demonstrations composed of state-action pairs. During inference, the robot can execute a previously unseen task by inferring the demonstrator's intent from only a few prompt demonstrations, without requiring any additional training.

Despite recent progress, existing robotic in-context imitation learning methods rely on state–action trajectories alone. By conditioning only on robot states (i.e., proprioceptive information and camera observations) and low-level actions, these approaches lack access to the underlying reasoning process that motivates the demonstrator’s decisions. This omission becomes particularly problematic in complex and ambiguous task settings, such as environments with many objects and multiple plausible task objectives, where the same actions may be consistent with different intents. In such scenarios, we hypothesize that explicit reasoning is crucial for conveying high-level task intent and guiding the learning process beyond surface-level in-context action imitation.

\begin{figure}
    \centering
    \includegraphics[width=\linewidth]{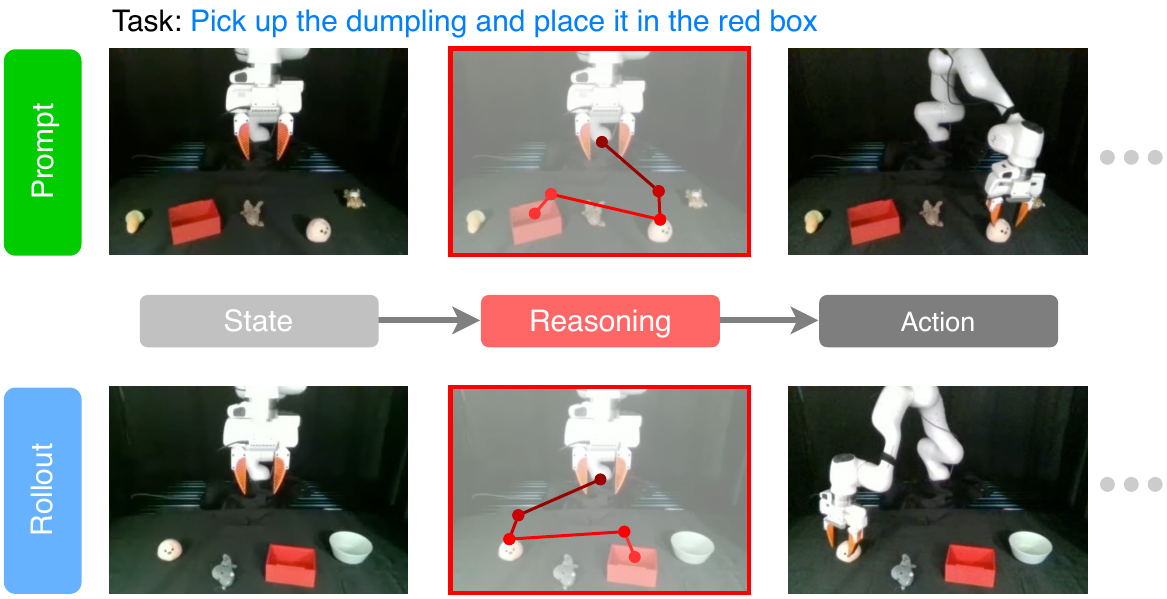}
    \caption{\textbf{General framework overview.} Our method augments prompt demos with keypoint-based visual reasoning traces in the image space, shown above with the overlaid polyline in the middle column. During inference, the model also performs visual reasoning before predicting the subsequent low-level robot action. The task's language description is included for clarity.}
    \label{fig:intro}
    % \vspace{-4ex}
\end{figure}

In this work, taking inspiration from advances in chain-of-thought prompting for large language models (LLMs) and large vision-language models (VLMs)~\cite{wei2022chain,zheng2023ddcot,zhang2024multimodal}, we propose \textbf{I}n-\textbf{C}ontext imitation \textbf{L}earning with visual \textbf{R}easoning (\method), a transformer-based method incorporating embodied visual reasoning into robotic in-context imitation learning. Specifically, our approach augments prompt demonstrations with explicit visual reasoning traces, in addition to the robot’s states and actions. These reasoning traces represent envisioned future robot trajectories in image space, capturing high-level task intent and providing structured guidance for action prediction. To execute a target task, conditioned on the augmented prompt demos, our method also generates the high-level visual reasoning traces before predicting the low-level control actions in an autoregressive manner. By learning to mimic not only robot actions but also the reasoning process underlying them, \method effectively grounds action prediction in structured task intent, leading to reliable adaptation to unseen tasks and strong generalization across visually complex scenarios, as validated via extensive experiments in both simulation and real-world settings. Figure~\ref{fig:intro} presents an overview of our method on the unseen task of putting the dumpling in the red box, given a single reasoning-augmented prompt demonstration.

In summary, our main contributions include:
\begin{itemize}
    \item We introduce \method, a novel in-context imitation learning method incorporating explicit embodied visual reasoning into demonstration prompts and policy inference.
    % \item We validate the effectiveness of our method via extensive experiments in both simulation and real-world robotic settings. 
    \item We evaluate our method through extensive experiments in both simulation and real-world robotic settings, demonstrating consistent performance improvements over competitive baselines and ablations.
\end{itemize}

% \begin{figure}
%     \centering
%     \includegraphics[width=\linewidth]{images/intro_3.pdf}
%     \caption{\textbf{General framework overview.} Our method augments prompt demonstrations with keypoint-based reasoning traces in the image space. In inference, the model also performs visual reasoning before predicting the subsequent low-level robot action.}
%     \label{fig:intro}
% \end{figure}

\section{RELATED WORK}

\subsection{In-Context Imitation Learning}
In-context imitation learning has recently emerged as a promising approach for enabling robots to adapt to new tasks from a small number of test-time demonstrations without further training~\cite{vosyliusinstant,vid2robot2024,fu2025icrt,sridhar2025ricl,Jiang2026RoboTTTCS}. A common recipe in existing in-context imitation learning methods is to construct prompt demos using only states or state-action pairs, and to directly predict actions for a target task.
For instance, ICRT~\cite{fu2025icrt} treats in-context robot learning as a next-token prediction problem and introduces an autoregressive framework for robotic in-context learning conditioning on teleoperated prompt demos consisting of the robot's proprioception, camera images, and robot actions.
% For example, KAT~\cite{di2024keypoint} represents sequences of observations and actions as sequences of keypoint-based tokens, converts them into text format, and leverages the in-context learning ability of an off-the-shelf LLM~\cite{achiam2023gpt}.
Another example is Vid2Robot~\cite{vid2robot2024}, which presents an encoder-decoder transformer that generates robot actions conditioned on the encoded representation of a human demo. While straightforward, this state-action formulation limits the robot's adaptability in cluttered and ambiguous environments, where the same actions may be consistent with different underlying task objectives and successful execution requires reasoning about task intent rather than direct action matching.

\subsection{Robotic Embodied Reasoning}
Recent work has explored incorporating reasoning processes into policy learning to improve robot navigation and manipulation in complex environments. Inspired by chain-of-thought reasoning in LLMs and VLMs~\cite{wei2022chain,zheng2023ddcot,zhang2024multimodal}, several approaches have investigated decomposing robotic tasks into intermediate reasoning steps, such as subgoals, plans, or structured representations, before executing low-level actions~\cite{huangthinkact,zawalski2025robotic,zhao2025cot,lee2025molmoact,team2025gemini,intelligence2026pi,fang2026molmoact2}. Among different embodied reasoning representations, visual reasoning, such as predicted future end-effector trajectories in image space~\cite{lee2025molmoact,lihamster2025,li2025coa,zhang2026peek}, offers a promising alternative to other language-based representations, which can be ambiguous and less compatible with continuous robot actions. Applying embodied reasoning to in-context imitation learning, however, remains an under-explored research problem. In this work, alongside proposing a novel in-context imitation learning method with embodied visual reasoning, we systematically benchmark multiple strategies for embodied reasoning integration. The results show that our proposed method achieves the strongest performance, leading to substantially improved adaptation to unseen and complex manipulation tasks across both simulation and real-world environments.

% \subsection{CoT Prompting}
% \lipsum[1]

\section{PROBLEM STATEMENT}% AND ASSUMPTIONS}
Our in-context imitation learning setting follows~\cite{fu2025icrt}. In particular, we consider a single-arm robot equipped with a standard gripper. There are two cameras: a third-view camera and a wrist-mounted camera. The goal is to train an in-context imitation learning method that can perform unseen tasks in novel environment configurations by conditioning on a few prompt demonstrations \textit{without any further training.}
Although the rollout environments allow the intended tasks demonstrated in the prompts, their configurations differ from those in the prompt demos, preventing the robot from naively copying the actions. Moreover, each testing configuration permits multiple possible tasks in addition to the desired one, requiring the model to infer the correct task objective from the prompt demos rather than relying on environmental cues alone.
% Note that the environment configurations of prompt demonstrations are different from those of the target tasks. 
A prompt demo typically consists of RGB observations captured from two cameras, robot proprioception, and robot action. In our method and other baseline methods that use visual reasoning traces, we augment the prompt demos by incorporating visual reasoning traces generated from third-view camera images as detailed in Section~\ref{subsec:gen_traces}.

% \begin{figure}[h]
%     \centering
%     \includegraphics[width=\linewidth]{example-image-duck}
%     \caption{Visual trace generation illustration caption}
%     \label{fig:trace_illustration}
% \end{figure}

\section{METHOD}\label{sec:method}

\begin{figure*}[t]
    \centering
    \includegraphics[width=\linewidth]{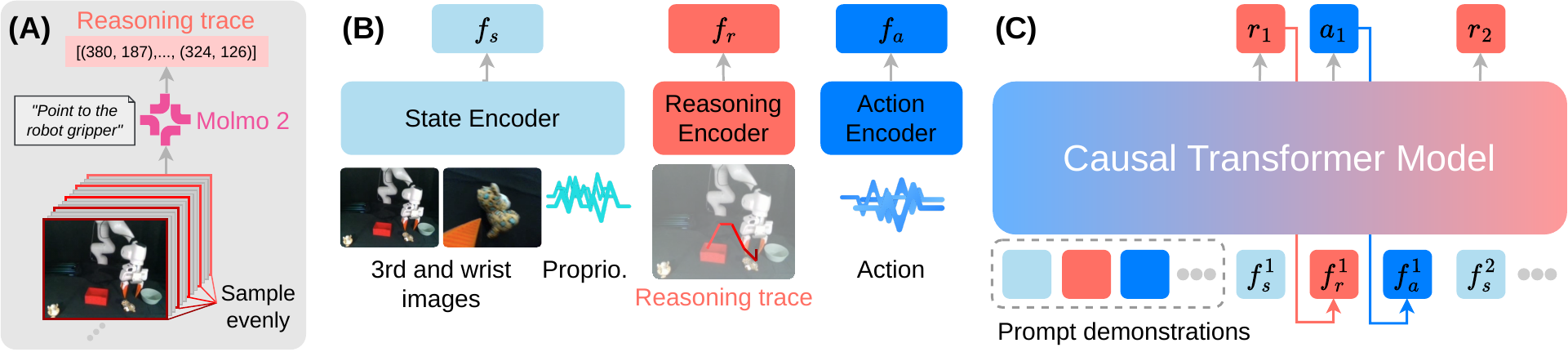}
    \caption{\textbf{Method overview.} \textbf{(A)} To generate the visual reasoning trace at a given time step, we uniformly sample five third-view images from that time step to the end of the trajectory and use Molmo2 to predict the gripper’s pixel location in each image. \textbf{(B)} Multi-view camera observations and proprioceptive states are encoded by a state encoder to produce state tokens $f_s$. Visual reasoning traces are embedded by a reasoning encoder to produce reasoning tokens $f_r$, and actions are embedded by an action encoder to produce action tokens $f_a$. \textbf{(C)} These modality-specific tokens are interleaved and fed into a causal transformer, which autoregressively predicts the next reasoning trace followed by the corresponding action. During training, teacher forcing is applied over reasoning and action tokens. In inference, the model first generates a reasoning trace and then produces the action in a closed-loop manner.
    % \yue{can we make the left part of the figure smaller? I feel it doesn't introduce too much information}}
    }
    \label{fig:method}
    % \vspace{-3ex}
\end{figure*}

In this section, we first outline the data formulation of in-context imitation learning. We then present the model architecture and implementation details of our \method method.

\subsection{Training Data Formulation}
For training in-context imitation learning policies, we consider a dataset $\mathcal{T}$ of visuomotor robot trajectories. In this dataset, each trajectory $\mathbf{T}^{\{i\}}=\left \{ \mathbf{s}_1, \mathbf{a}_1,..., \mathbf{s}_{t_i}, \mathbf{a}_{t_i}\right \}$ is a ${t_i}$-step sequence of states $\mathbf{s}$ (including multi-view camera observations and the robot proprioceptive information), and robot actions $\mathbf{a}$. We use the absolute end-effector pose as the robot's proprioception and the delta end-effector pose between consecutive time steps as the control action. To facilitate in-context learning, the dataset $\mathcal{T}$ is split into $K$ disjoint subsets $\mathcal{T} = \bigcup_{k=1}^{K}\mathcal{S}_k$, $\mathcal{S}_k \cap \mathcal{S}_l = \emptyset$ for $k\neq l$. Each $\mathcal{S}_k$ is a subset of trajectories of a single task, as determined by the semantic labels associated with the trajectories. These labels are typically textual descriptions of the performed tasks, such as \textit{``Poke the lion.''} A training sequence is the concatenation of trajectories of the same subset, where the first randomly-selected $n$ trajectories serve as prompt demos and the remaining trajectories are treated as target episodes.

\subsection{Visual Reasoning Trace Generation}\label{subsec:gen_traces}
To incorporate visual reasoning into in-context imitation learning, we first augment each trajectory $\mathbf{T}^{\{i\}}$ of $\mathcal{T}$ with visual reasoning traces $\mathbf{r}$ generated from the third-view RGB observations in $\mathbf{o}$, forming the reasoning-augmented dataset $\mathcal{T}_{\text{aug}}$ of trajectories $\mathbf{T}^{\{i\}}_{\text{aug}}=\left \{  \mathbf{s}_1, \mathbf{r}_1, \mathbf{a}_1,..., \mathbf{s}_{t_i}, \mathbf{r}_{t_i}, \mathbf{a}_{t_i}\right \}$. The format of our visual reasoning traces follows MolmoAct~\cite{lee2025molmoact}. In particular, at each time step in $\mathbf{T}^{\{i\}}_{\text{aug}}$, the visual reasoning trace is a polyline of 5 points corresponding to the robot gripper's positions in the pixel space of third-view observations, sampled evenly from the future horizon of the episode between the current and the terminal time step. We choose five points as a practical balance between granularity and efficiency for our experiments, which largely involve pick-and-place tasks (see Section~\ref{sec:exps}). In this setting, the five points naturally align with the four key stages of the behavior: moving to the target object, grasping it, transporting it to the receptacle, and placing the object. For longer-horizon tasks, the visual trace format can be designed to include more points to provide finer temporal resolution. Note that, in contrast to MolmoAct, where visual traces are represented in textual form, we represent visual traces as numerical vectors. To determine the gripper's position in an image, in simulation, we utilize the gripper's 3D position (inferred from the robot's proprioceptive state) and the known camera parameters. In the real world, since the camera parameters are usually unavailable, we employ the Molmo2 VLM~\cite{clark2026molmo2} and prompt it with the command \textit{``Point to the robot gripper.''} In our work, we found that the gripper positions detected by Molmo2 are of high precision and facilitate our imitation learning with visual reasoning. Nevertheless, the generation of visual reasoning traces following this approach is not tied to a specific model and can be implemented using a wide range of VLMs~\cite{team2025geminier, Bai2025Qwen3VLTR}, segmentation models~\cite{ravisam,simeoni2025dinov3}, or detection models~\cite{carion2020end,minderer2023scaling}. In our work, we choose Molmo2 because it is open-source and has demonstrated its state-of-the-art performance on pointing tasks and usefulness for several downstream robotic applications~\cite{lee2025molmoact,hong2025hand,chen2026steerable}.
% An illustration of visual trace generation is shown in Figure~\ref{fig:trace_illustration}.

\subsection{In-Context Imitation Learning with Visual Reasoning}\label{subsec:ICLR}
In the following, we describe the implementation details, training, and inference processes of our \method method.

\textbf{Model Architecture.} We adopt a Llama2-style~\cite{touvron2023llama} causal transformer architecture similar to~\cite{fu2025icrt}, with modality-specific encoders for states, reasoning traces, and actions. Robot states are encoded using a state encoder. In particular, visual observations from the third-view and wrist cameras are first encoded using a pretrained vision transformer~\cite{fu2025icrt}, while proprioceptive information is embedded by an MLP. We then employ attention pooling~\cite{lee2019set} to aggregate visual patch tokens and proprioceptive features to form the state token $f_s$. Visual traces, which are represented as ordered sets of keypoints, are flattened and encoded using an MLP reasoning encoder, producing reasoning tokens $f_r$. Similar to proprioception, the action tokens $f_a$ are encoded by another MLP action encoder. All modality embeddings are interleaved into a single token sequence and processed by the transformer using next-token prediction. The overview of our model architecture is illustrated in Figure~\ref{fig:method}. Compared to ICRT~\cite{fu2025icrt}, the key architectural difference lies in the inclusion of visual reasoning tokens $f_r$ in the input and output sequence, enabling the transformer to jointly model reasoning and action generation within a unified autoregressive framework.

\textbf{Loss Functions.} We employ the standard next-token prediction objective, with losses applied only to predictions after the prompt demonstrations to preserve the in-context learning behavior. The loss is computed over both reasoning trace prediction and action prediction for target episodes. In particular, the combined loss is computed as
\begin{equation}
\mathcal{L} = \mathcal{L}_{\text{action}} + 0.3 \times \mathcal{L}_{\text{reasoning}},
\end{equation}
where $\mathcal{L}_{\text{action}}$ and $\mathcal{L}_{\text{reasoning}}$ are L1 losses for action prediction and reasoning trace prediction, respectively. We set the reasoning loss weight to 0.3, which empirically achieves a balanced trade-off between action and reasoning learning. 

\textbf{Training.} During training, we freeze the pretrained vision transformer encoding camera images. We also apply action chunking~\cite{zhao2023learning}, where the model is trained to predict the next 16 actions instead of a single one. In addition, we randomly mask a subset of the visual reasoning trace tokens in the target trajectories, while preserving all reasoning trace tokens in the prompt demonstrations. In particular, for each training sequence, we sample a random masking ratio from 0\% to 100\% and uniformly mask the corresponding number of reasoning tokens in the target portion of the sequence. This masking acts as a regularization technique that prevents the model's action prediction from over-relying on its generated reasoning traces, which are highly correlated with the actions but may be imperfect in challenging settings. As a result, the model learns to remain robust when reasoning traces are noisy or partially missing. This design, therefore, also naturally enables an efficient inference-time variant, which we refer to as reasoning dropout, inspired by similar strategies in~\cite{chen2025training,intelligence2026pi}, where the model omits reasoning trace generation during inference despite being trained to generate them. We adopt a teacher-forcing training scheme, where ground-truth reasoning and action tokens are used instead of the model’s own predictions when conditioning the next-token prediction.

\textbf{Inference.} At test time, the human demonstrator provides one or more teleoperated demos consisting of state–action pairs. Molmo2 is used for detecting the robot gripper's positions in the third-view camera images concurrently with the teleoperation session. After an episode is completely recorded, the visual reasoning traces for that episode are generated following the visual trace generation process described in Section~\ref{subsec:gen_traces}. Conditioned on the augmented prompt demonstrations and the current state, the model first predicts the next visual reasoning trace and then the corresponding action chunk, of which the executed immediate action is computed via temporal ensembling~\cite{zhao2023learning}. After executing each action, the policy receives the updated state, enabling it to iteratively generate the next reasoning trace before predicting and executing subsequent actions. We apply key-value caching~\cite{touvron2023llama} to accelerate our transformer model's inference, where previously computed transformer key and value states are reused, enabling incremental decoding without recomputing the full sequence. Practically, it takes our model roughly \SI{0.0278}{\milli\second} to predict a visual trace or an action on an NVIDIA GeForce RTX 5090 GPU. For the reasoning dropout variant of our method, the visual trace generation step is skipped, and a zero vector is used in place of the reasoning trace to condition action prediction.

\section{EXPERIMENTS}\label{sec:exps}

\begin{table}[t]
\centering
% \vspace{1ex}
% \setlength{\tabcolsep}{3pt}
\resizebox{\linewidth}{!}{%
\begin{tabular}{lccccc}
\toprule
& & \multicolumn{3}{c}{\bf{LIBERO-90}} & \\
 \cmidrule(lr){3-5}
\bf{Method}
& \bf{LIBERO-Object} & \bf{Kitchen} & \bf{Living} & \bf{Study} & \cellcolor{myorange}\bf{Avg.} \\
\midrule
ICRT~\cite{fu2025icrt}
& 44.44 & 0.89 & \underline{18.93} & 0.83 & \cellcolor{myorange}16.27\\
TO Dropout
& 62.22 & 17.56 & 16.67 & 25.00 & \cellcolor{myorange}30.36\\
TO
& 54.44 & 12.11 & 12.80 & 29.33 & \cellcolor{myorange}27.17\\
\midrule
Ours Dropout
& \bf{70.89} & \bf{60.22} & \bf{38.93} & \bf{46.17} & \cellcolor{myorange}\bf{54.05} \\
Ours
& \underline{70.00} & \underline{20.00} & 11.20 & \underline{32.17} & \cellcolor{myorange}\underline{33.34}\\
\bottomrule
\end{tabular}
}
% \vspace{-0.5ex}
\caption{\textbf{Simulation success rates (\%).} ``Dropout'' models are models that learn to generate visual reasoning traces in training but skip the reasoning steps for the target trajectory in inference. Results are reported on 3 unseen tasks of LIBERO-Object and 15 unseen tasks of LIBERO-90 (6 of kitchen scenes, 5 of living room scenes, and 4 of study scenes). In each column, the highest score is \textbf{bolded}, and the second-best performance is \underline{underlined}.}
\label{tab:sim_results}
% \vspace{-1ex}
\end{table}

Here, we detail our experimental setups and results in both simulation and real-world settings to evaluate our \method.

\subsection{Models}
 As mentioned in Section~\ref{subsec:ICLR}, our training mechanism allows for two model variants at test time, one complete model and one that does not predict visual traces for the target trajectory (while still conditioning on prompt demos augmented with visual traces). We include both of them in our experiments and denote them as \textbf{Ours} and \textbf{Ours Dropout}. Additionally, we employ ICRT~\cite{fu2025icrt}, a state-of-the-art in-context imitation learning framework, as a baseline to compare with our models. As described earlier, our approach builds directly upon the ICRT architecture. Comparing against ICRT allows us to isolate and evaluate the effect of integrating visual reasoning into the in-context imitation learning framework. Although we do not compare our method with certain prior approaches due to differences in experimental settings, for example, Instant Policy~\cite{vosyliusinstant} relies on two external depth cameras, while KAT~\cite{di2024keypoint} requires a calibrated wrist-mounted camera, we believe that the principle of visual reasoning proposed is broadly applicable and can be adapted to other robotic setups. To further evaluate the importance of including visual reasoning traces in prompt demonstrations, we implement another target-only (TO) reasoning approach in which visual reasoning traces are omitted from the prompt demonstrations, while the model is still trained to generate reasoning traces before predicting actions for the target trajectories. We apply the same reasoning dropout training strategy used in our method to this method. As a result, we obtain two versions of this approach for inference, denoted as \textbf{TO} and \textbf{TO Dropout} in our experimental results. To ensure fair comparisons, any components shared between models are kept identical. We train all models on 2 NVIDIA A6000 Ada GPUs for both simulation and real-world experiments.

\subsection{Simulation Experiments}\label{subsec:sim_exp}
\textbf{Setup.} We first perform our experiments in LIBERO~\cite{liu2023libero}, a widely-used simulation benchmark for robot learning, with a Franka Panda robot arm. In particular, we use the two LIBERO-Object and LIBERO-90 task suites. There is no standard setting for in-context imitation learning on LIBERO, so we repurpose it for our experiments. More specifically, for LIBERO-Object, which has 10 tasks, we randomly select 7 tasks for training, and reserve the remaining 3 as unseen tasks for testing. In LIBERO-90, there are 90 tasks that span 3 different environments (i.e., kitchen scenes, living room scenes, and study scenes). We apply a train/test ratio of 75/15, ensuring that both the training and testing sets contain tasks from all three environments. In particular, we randomly select 6 testing tasks for kitchen scenes, 5 for living room scenes, and 4 for study scenes. Each task in LIBERO comes with 50 expert demos, all of which are used for training. Following~\cite{kim2025openvla}, we apply a preprocessing step to filter out no-op actions and unsuccessful demonstrations. We inherit training hyperparameters from~\cite{fu2025icrt} for all models. During testing, for each unseen task, we use three different prompt episodes (expert demonstrations of that task) and evaluate the model under 50 distinct task initializations with varying object configurations. This results in 150 rollouts per task per model. We report the overall average success rate across all tasks for LIBERO-Object and the average success rate for each environment for LIBERO-90. We also report the overall success rate aggregated across all four settings of LIBERO.

\begin{table*}[t]
\centering
% \caption{Real-world task success rates (\%) \yue{Are there any other baselines?}}
% \vspace{1ex}
\setlength{\tabcolsep}{3pt}
\resizebox{\textwidth}{!}{%
\begin{tabular}{lcccccccccccccc}
\toprule
& \multicolumn{7}{c}{\bf{Poking}} & \multicolumn{7}{c}{\bf{Pick-and-Place}} \\
\cmidrule(lr){2-8} \cmidrule(lr){9-15}
\bf{Method}
& \bf{Hippo} & \bf{Dumpling} & \bf{Jaguar} & \bf{Monkey} & \bf{Lion} & \bf{Potato} & \cellcolor{myorange}\bf{Avg.}
& \bf{\makecell{Dumpling to \\ Red box}} & \bf{\makecell{Zebra to \\ Blue bowl}} & \bf{\makecell{Tomato to \\ Grey bowl}} & \bf{\makecell{Monkey to \\ Red box}} & \bf{\makecell{Lion to \\ Blue bowl}} & \bf{\makecell{Potato to \\ Grey bowl}} & \cellcolor{myorange}\bf{Avg.} \\
\midrule
ICRT~\cite{fu2025icrt}
& 20 & 40 & 80 & 60 & 40 & 50 & \cellcolor{myorange}48.33
& 15 & 35 & 30 & 15 & 25 & 5 & \cellcolor{myorange}22.50 \\

TO Dropout
& 0 & 60 & 40 & 40 & 60 & 20 & \cellcolor{myorange}40.00
& 55 & 50 & 30 & 5 & 40 & 10 & \cellcolor{myorange}31.67 \\

TO
& 30 & 60 & 80 & 30 & 40 & 70 & \cellcolor{myorange}51.67
& 65 & 45 & 45 & 55 & 20 & 15 & \cellcolor{myorange}40.83 \\

\midrule

Ours Dropout
& 70 & 60 & 60 & 60 & 70 & 70 & \cellcolor{myorange}\underline{65.00}
& 55 & 55 & 50 & 45 & 30 & 45 & \cellcolor{myorange}\underline{46.67} \\

Ours
& 60 & 80 & 80 & 70 & 70 & 70 & \cellcolor{myorange}\bf{71.67}
& 65 & 70 & 65 & 50 & 45 & 65 & \cellcolor{myorange}\bf{60.00} \\
\bottomrule
\end{tabular}
}
% \vspace{-0.5ex}
\caption{\textbf{Real-world task success rates (\%).} Every model has 10 rollouts for each task. See Section~\ref{subsec:real_exp} for our detailed evaluation methodology.
% \yue{Are there any other baselines?}
}
\label{tab:real_world_results}
% \vspace{-3ex}
\end{table*}

\textbf{Results.} The results are shown in Table~\ref{tab:sim_results}, indicating that our models (both the complete and dropout variants) significantly outperform other baselines. In particular, our dropout model consistently obtains the highest success rates across all settings, while the complete \method model achieves the second-best scores on three out of four settings and on the overall average success rate. The target-only models rank third and fourth, while ICRT performs the worst overall. These results demonstrate the effectiveness of incorporating explicit reasoning into in-context imitation learning and underscore the importance of including visual reasoning traces in prompt demos, which enable the model to learn to reason and, in turn, facilitate better action prediction in unseen tasks.

\begin{figure}
    \centering
    \includegraphics[width=\linewidth]{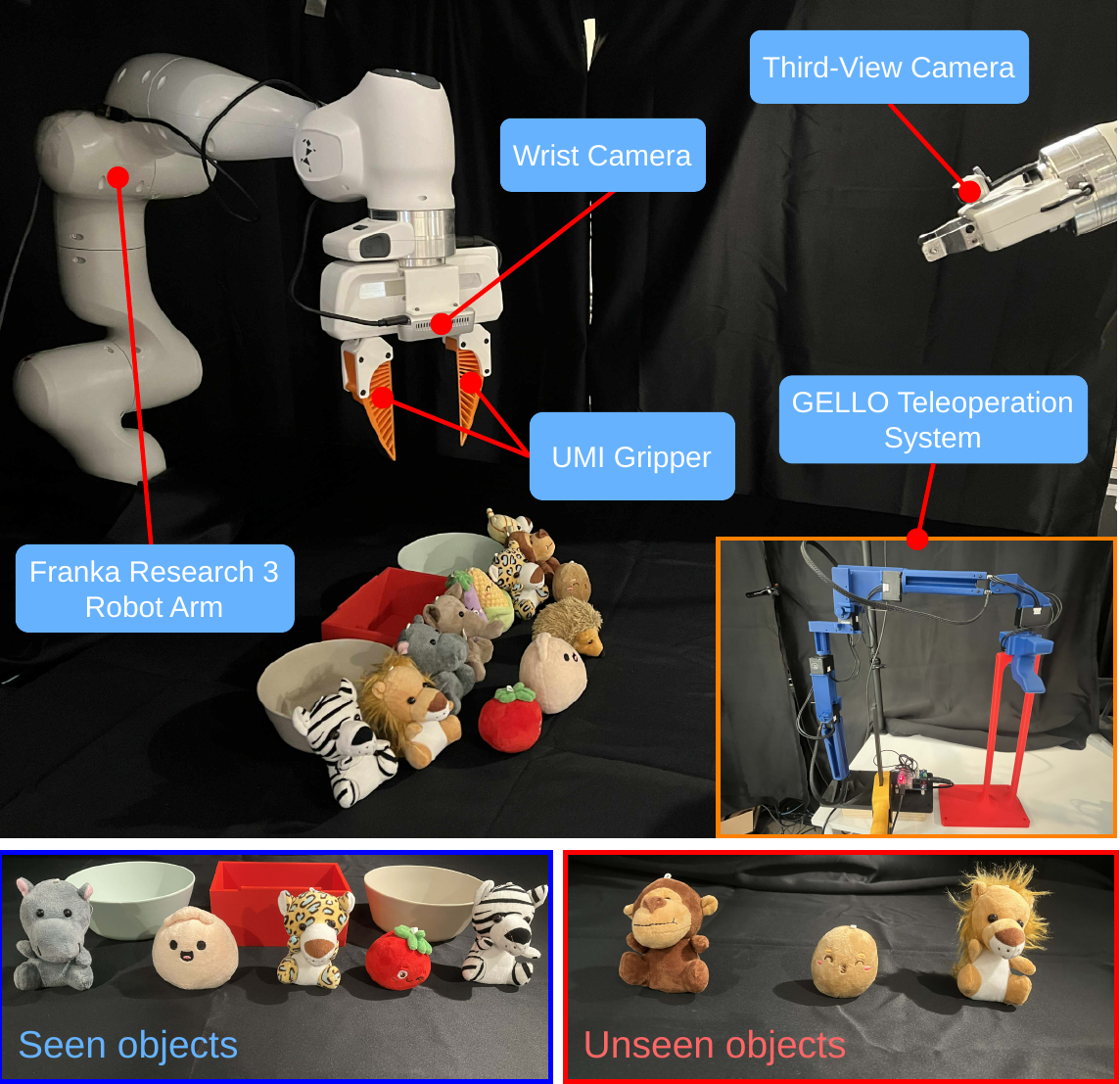}
    \caption{\textbf{Real robot setting.} We use a Franka Research 3 robot arm equipped with a UMI gripper. Visual observations are captured by two RealSense cameras. Teleoperation for data collection and test-time prompt demonstration recording is performed using a GELLO system. Testing objects appearing in training episodes are shown in the bottom-left box, while completely unseen testing objects are shown in the bottom-right box.}
    \label{fig:real_setting}
    % \vspace{-3ex}
\end{figure}

\subsection{Real Robot Experiments}\label{subsec:real_exp}

\begin{figure*}[t]
    \centering
    \includegraphics[width=\linewidth]{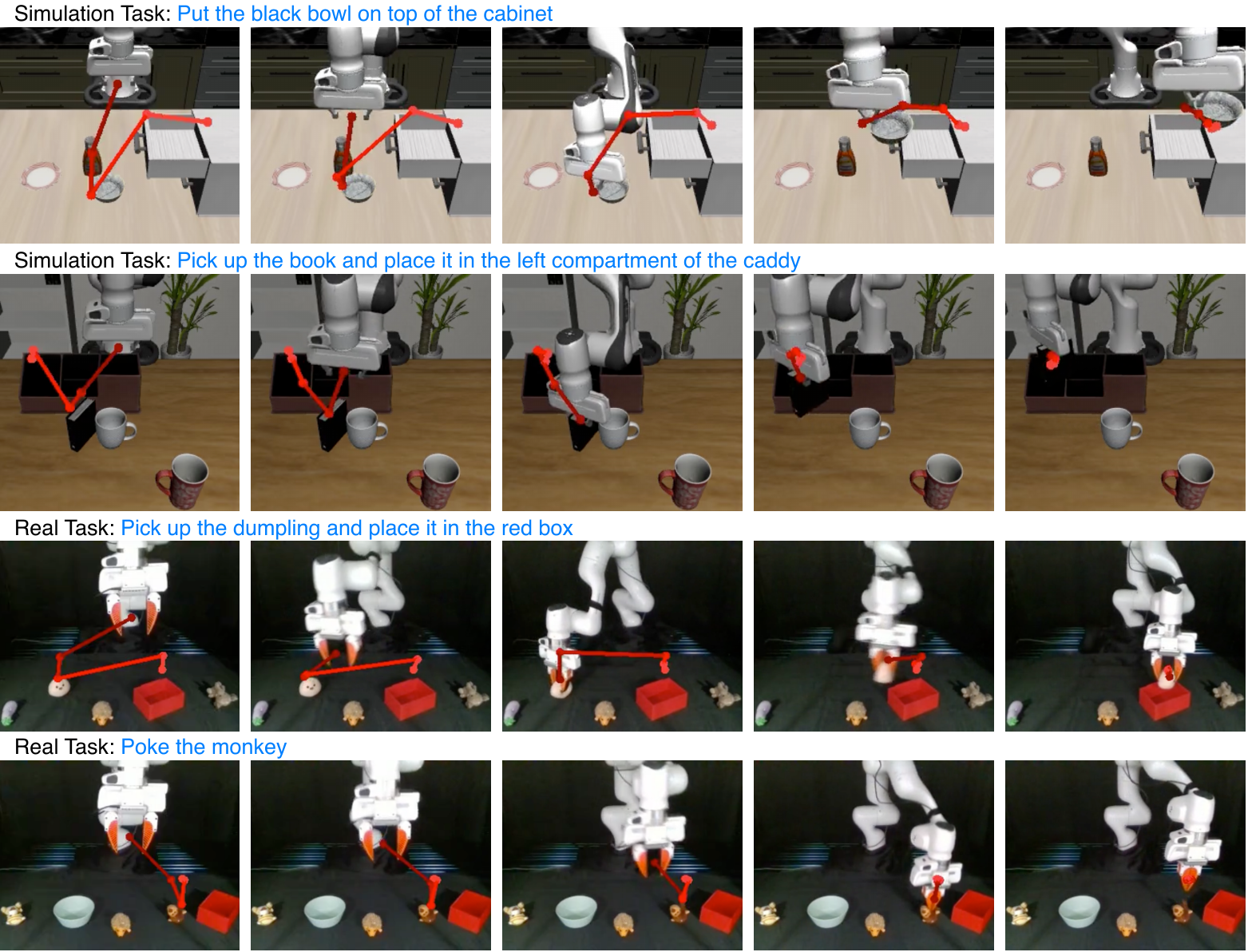}
    \caption{\textbf{Qualitative results.} Rollout examples of our complete \method model in simulation (first two rows) and real-world settings (two bottom rows). All presented visual traces are predicted by our model.}
    \label{fig:qualitative_results}
    % \vspace{-2ex}
\end{figure*}

\textbf{Setup.} In the real world, we consider a tabletop manipulation setup using a Franka Research 3 robot arm with a UMI gripper~\cite{chi2024universal}. We use the UMI gripper because of its compliance and its bright color, which facilitates more reliable detection by Molmo2. We set up two cameras, one RealSense D435if wrist-mounted camera and another RealSense D415 camera as the fixed third-view camera. To collect training data and record prompt trajectories at test time, we employ a GELLO teleoperation system~\cite{wu2024gello}. Our real-world setting is depicted in Figure~\ref{fig:real_setting}. For training, we collect 825 demonstrations for 5 poking and 10 pick-and-place tasks across many objects. In each training demo, we ensure that multiple tasks are feasible from the same initial configuration. This prevents shortcut learning and instead encourages the models to perform in-context learning by inferring task intent from the prompt demonstrations. We deploy all models at 30Hz using a computer with a 32GB NVIDIA GeForce RTX 5090 GPU connected to the robot.

\textbf{Evaluation Protocol.} The models are evaluated on 12 unseen tasks that are not included in the training data. Specifically, there are 6 unseen poking tasks, corresponding to the objects \textit{hippo}, \textit{dumpling}, \textit{jaguar}, \textit{monkey}, \textit{lion}, and \textit{potato}. Additionally, there are 6 unseen pick-and-place tasks, which are \textit{dumpling to red box}, \textit{zebra to blue bowl}, \textit{tomato to grey bowl}, \textit{monkey to red box}, \textit{lion to blue bowl}, and \textit{potato to grey bowl}. Note that \textit{hippo}, \textit{dumpling}, \textit{jaguar}, \textit{zebra}, \textit{tomato}, \textit{red box}, \textit{blue bowl}, and \textit{grey bowl} are objects appearing in training (although the associated tasks are unseen), while \textit{monkey}, \textit{lion}, and \textit{potato} are completely unseen objects. Following the evaluation setting of~\cite{fu2025icrt}, each task has five levels of difficulty, with the number of distractor objects increasing. For each model, we perform 10 rollouts per task, with two rollouts (different configurations) for each difficulty level. For each task, we record 3 prompt demonstrations, i.e., demos with zero, one distractor object, and a distractor receptacle for pick-and-place, or two distractor objects for poking tasks. For every rollout, a one-in-three random prompt demo is used. In a poking task rollout, a model receives 1 point if the object is poked by the robot gripper. In a pick-and-place rollout, the model is scored with 0.5 for a successful pick and 1 if the object is also placed in the correct receptacle. For each rollout, the model has 300 steps for retries. Note that each training episode has from 80 to 200 steps. We report the success rate for each task, as well as the average success rate across the two task types for all models.

\textbf{Results.} The results are shown in Table~\ref{tab:real_world_results}, where similar to simulation results, our models largely outperform other baselines on both poking and pick-and-place tasks, reaffirming the effectiveness of our proposed method. Interestingly, while the dropout variants (TO Dropout and Ours Dropout) achieve better performance than the full models in simulation, the complete models obtain higher success rates in real-world settings. We hypothesize that this discrepancy arises from differences in scene configurations between training and testing in simulation and real-world settings. In the LIBERO simulation, the differences between training and testing scene configurations, particularly object positions, are relatively small. This reduces the need for explicitly generating visual traces and allows the dropout models to ``internalize'' the reasoning process. We also observe that, in the LIBERO experiments, the actions generated by the dropout models are more stable than those produced by the full models. This may be due to the limited diversity of visual traces in the training data, which restricts the model’s ability to reliably learn trace generation. As a result, errors in predicted visual traces can propagate to action prediction, reducing the overall stability of the complete models' generated actions. Together, these factors contribute to the better performance of the dropout models in the LIBERO experiments. In contrast, real-world training data is substantially more diverse and the differences between training and testing configurations are considerably bigger than in LIERBO. Consequently, the models can learn to generate visual traces more effectively and explicit reasoning becomes much more important for guiding action prediction, leading to the superior performance of the full reasoning-enabled models. Refer to Figure~\ref{fig:qualitative_results} for rollout examples of our complete \method model. In fact, we find that the reasonably strong performance of our \method dropout variant, along with the discrepancy between low- and high-diversity settings, aligns with empirical observations from previous works on embodied reasoning~\cite{intelligence2026pi,chen2025training}.

\subsection{Ablation Studies}~\label{subsec:ablation_studies}
In this section, we conduct additional experiments in the real-world setting to further evaluate our method.

\begin{figure}[t]
    \centering
    \includegraphics[width=\linewidth]{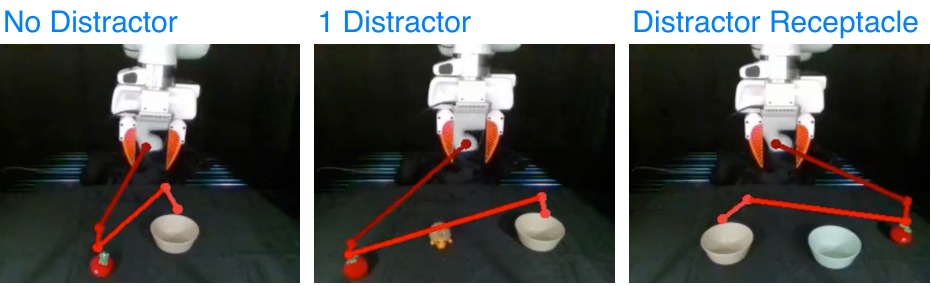}
    % \vspace{-3.5ex}
    \caption{\textbf{Three types of prompt demonstrations.} The task of picking up the tomato and putting it in the grey bowl is selected.}
    \label{fig:prompt_types}
    % \vspace{-2.5ex}
\end{figure}

\textbf{Prompt Demonstrations.} We investigate the effect of different types of prompt demos on the models' performances. We consider the task of putting the tomato in the grey bowl. Similar to the evaluation protocol described in Section~\ref{subsec:real_exp}, we collect three prompt demonstrations: one with no distractor, one with one distractor, and another with one distractor receptacle (see Figure~\ref{fig:prompt_types}). We evaluate the models using five prompt configurations: each of the three demos individually, a two-demo configuration where two demos are randomly sampled for every rollout, and a three-demo configuration containing all demonstrations. We conduct 10 rollouts per model for each prompt configuration and report the success rates in Table~\ref{tab:prompt_trajectories}. The results indicate that our complete \method model consistently achieves the highest success rates, demonstrating its stability under different prompt types. We hypothesize that this is attributable to the diverse range of prompt demonstrations the model encounters during training. We also observe that, across all models, increasing the number of prompt demos does not necessarily lead to a clear improvement in performance, aligning with similar findings in other in-context robot learning works~\cite{fu2025icrt,di2024keypoint,ruoss2025lmact}. This contrasts with the typical behavior observed in LLMs and VLMs, where performance generally improves as the number of in-context examples increases~\cite{brown2020language}. We leave a deeper investigation of this observation to future work.

\begin{table}[t]
\centering
% \caption{Results of different prompt types (\%)}
% \vspace{1ex}
% \setlength{\tabcolsep}{3pt}
\resizebox{\linewidth}{!}{%
\begin{tabular}{lccccc}
\toprule
& \multicolumn{3}{c}{\bf{1 Demo}} & & \\
 \cmidrule(lr){2-4}
\bf{Method}
& \bf{0 Distr.} & \bf{1 Distr.} & \bf{Distr. Receptacle} & \bf{2 Demos} & \bf{3 Demos} \\
\midrule
ICRT~\cite{fu2025icrt}
& 35 & 30 & 40 & 35 & \underline{45} \\
TO Dropout
& \underline{45} & 25 & 20 & 25 & \underline{45} \\
TO
& {40} & \underline{40} & \underline{50} & {50} & 30 \\
\midrule
Ours Dropout
& 35 & 35 & 40 & \underline{55} & 40 \\
Ours
& \bf{55} & \bf{65} & \bf{55} & \bf{70} & \bf{65} \\
\bottomrule
\end{tabular}
}
% \vspace{-0.5ex}
\caption{\textbf{Results of different prompt types (\%).} ``Distr.'' stands for distractor. Success rates are calculated over 10 trials for each experiment on the task of picking up the tomato and putting it in the grey bowl.}
\label{tab:prompt_trajectories}
% \vspace{-2ex}
\end{table}

\begin{figure}[t]
    \centering
    \includegraphics[width=\linewidth]{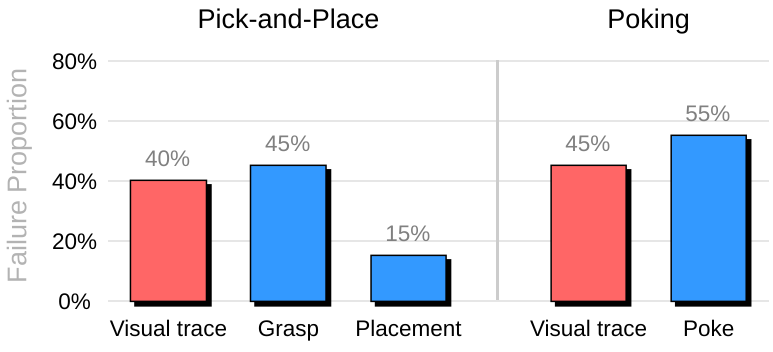}
    % \vspace{-4ex}
    \caption{\textbf{Failure analysis.} We report the proportion of visual trace errors (\textcolor{myred}{red}) versus other failure types (\textcolor{myblue}{blue}), computed over 20 failed rollouts for each of the pick-and-place (tomato to grey bowl) and poking (hippo) tasks.}
    \label{fig:failure_analysis}
    % \vspace{-3ex}
\end{figure}

\textbf{Failure Analysis.} The incorporation of reasoning improves our models' transparency, as the generated visual traces provide interpretable intermediate representations that help us better understand the robot's behavior. Leveraging this interpretability, we systematically analyze the relative proportions of different failure causes in inference. We consider the pick-and-place task of putting the tomato in the grey bowl and the task of poking the hippo. For the pick-and-place task, we categorize failures into three types. The first is visual trace errors, where the generated traces target the wrong object to grasp or terminate at an incorrect receptacle. If the visual traces are correct, failures are further classified as either grasp failures (i.e., failing to grasp the tomato) or placement failures (i.e., failing to place the grasped tomato into the grey bowl). For poking, failures are divided into visual trace errors (traces point to the wrong object) and poking failures (failure to poke the hippo given correct reasoning traces). For each task, we report the percentages over 20 failed rollouts, as shown in Figure~\ref{fig:failure_analysis}. In the pick-and-place task, grasp failure is the most significant error type, while in the poking task, mis-poking accounts for the largest proportion. Although visual trace errors contribute 40\% and 45\% of failures in pick-and-place and poking, respectively, they are not the primary failure mode in either task. This suggests that the proposed integration of visual reasoning is generally effective at capturing task intent, and that overall performance is more often limited by downstream execution challenges rather than incorrect reasoning. Moreover, visual trace errors could potentially be mitigated by improved gripper localization models (e.g., future Molmo versions), which would likely translate into additional performance gains. Overall, the results indicate that improving low-level control robustness may further enhance performance, building upon the strong reasoning capability of our method.

\begin{figure}[t]
    \centering
    \includegraphics[width=\linewidth]{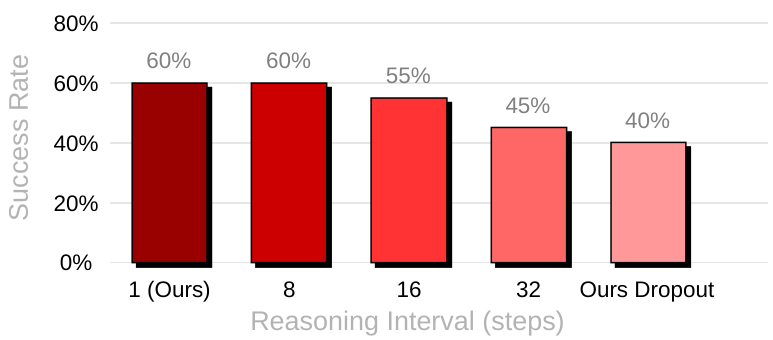}
    % \vspace{-2ex}
    \caption{\textbf{Results of different reasoning intervals.} The experiment is conducted on the task of putting the hedgehog into the red box, with 10 rollouts for each model variant.}
    \label{fig:efficient_reasoning}
    % \vspace{-2ex}
\end{figure}

\textbf{Efficient Reasoning.} As the proposed reasoning-dropout training approach allows our model to either predict the reasoning trace or omit it at any time step during inference, we further conduct an experiment to benchmark the performance of different model variants with varying reasoning intervals. In particular, in addition to our complete \method model and the dropout model (which completely ignores reasoning during testing), we evaluate three additional reasoning-efficient variants that perform visual reasoning every 8, 16, and 32 time steps. For each model, we perform 10 rollouts on the unseen task of picking up the hedgehog and placing it in the red box. The corresponding success rate results are shown in Figure~\ref{fig:efficient_reasoning}. In general, we observe a decreasing trend in performance as the interval between consecutive reasoning steps increases, with the complete model achieving the highest performance and the dropout model performing the worst. However, the 8-step and 16-step variants achieve results comparable to the full \method model, with the 8-step variant performing on par with the complete model while being roughly 8$\times$ faster. The results suggest that while explicit test-time reasoning is important for our method, we can reduce, to some extent, the frequency with which the model performs reasoning while still achieving commendable performance.

% \textbf{Reasoning Steerability.}~\lipsum[1]
% \begin{figure}[t]
%     \centering
%     \includegraphics[height=4cm, width=\linewidth]{example-image-duck}
%     \caption{\textbf{Reasoning steerability.}}
%     \label{fig:steerability}
% \end{figure}

\section{DISCUSSION}
Despite promising results, our work still has considerable room for improvement. Current experiments involve two types of tasks. Experimenting on a wider range of tasks would further validate the generalizability of our \method. While the current visual reasoning representation has demonstrated its usefulness, incorporating other forms of reasoning (e.g., bounding boxes, affordances, or depth information) is promising to benefit in-context imitation learning. Additionally, although in-context imitation learning aims to enable data-efficient robot policies, collecting training data remains time-consuming. A worthwhile yet under-investigated research direction is to develop in-context robot learning methods that can effectively condition on human video demos or demos collected on different robot embodiments. Looking further ahead, we believe that the paradigm of robotic in-context imitation learning remains underexplored. Designing in-context learning methods that can scale to bimanual, dexterous, and long-horizon manipulation is an important open research question that requires advances in both reasoning representations and hierarchical policy learning. We leave these promising research directions for future work.

\section{CONCLUSIONS}
We present \method, a novel and effective method that incorporates visual reasoning into robotic in-context imitation learning. We conduct a wide range of simulation and real-world robotic experiments, where our proposed method consistently outperforms other methods by a large margin, demonstrating its stronger generalization to unseen tasks and novel object configurations. These results suggest that embodied visual reasoning is a promising direction for improving the robustness and adaptability of robotic in-context imitation learning systems.

\section{ACKNOWLEDGMENTS}
We sincerely thank our friends and colleagues for their support throughout this project. In particular, we are grateful to Kyle Hatch, Nhat Chung, and Sicheng He for their insightful discussions and valuable suggestions. We also thank Sicheng He, Bo-Ruei Huang, and Jason Chen for their assistance in repairing the robot. This work is partially supported by the National Science Foundation through NSF CPS \#2434460. The USC Physical Superintelligence Lab acknowledges generous support from Toyota Research Institute, Dolby, Google DeepMind, Capital One, Nvidia, and Qualcomm. Yue Wang is also supported by a Powell Research Award.


%%%%%%%%%%%%%%%%%%%%%%%%%%%%%%%%%%%%%%%%%%%%%%%%%%%%%%%%%%%%%%%%%%%%%%%%%%%%%%%%

%%%%%%%%%%%%%%%%%%%%%%%%%%%%%%%%%%%%%%%%%%%%%%%%%%%%%%%%%%%%%%%%%%%%%%%%%%%%%%%%


% \section*{ACKNOWLEDGMENT}

%%%%%%%%%%%%%%%%%%%%%%%%%%%%%%%%%%%%%%%%%%%%%%%%%%%%%%%%%%%%%%%%%%%%%%%%%%%%%%%%

\balance
\bibliographystyle{IEEEtran}
\bibliography{reference}

\end{document}